\documentclass{article}

\usepackage{arxiv}
\usepackage[utf8]{inputenc} 
\usepackage[T1]{fontenc}     
\usepackage{hyperref}          
\usepackage{url}                    
\usepackage{booktabs}        
\usepackage{amsfonts}        
\usepackage{nicefrac}          
\usepackage{microtype}       
\usepackage{lipsum}
\usepackage{amsmath}
\usepackage{graphicx}
\usepackage{chngcntr}
\usepackage{cite}
\counterwithin{table}{section}

\title{Deep Learning Models to Predict Pediatric Asthma Emergency Department Visits}

\author{
Xiao Wang, PhD; Zhijie Wang; Yolande M. Pengetnze, MD; Vikas Chowdhry \\
Parkland Center for Clinical Innovation (PCCI) \\
Dallas, TX, USA \\  
\textit{xiao.wang@pccinnovation.org, zhijie.wang@pccinnovation.org} \\ \textit{yolande.pengetnze@pccinnovation.org, vikas.chowdhry@pccinnovation.org} \\
\AND
Barry S. Lachman, MD\\
Parkland Community Health Plan (PCHP) \\
Dallas, TX, USA \\ 
\textit{barry.lachman@phhs.org} \\
}
\date{}

\begin{document}
\maketitle

\begin{abstract}

Pediatric asthma is the most prevalent chronic childhood illness, afflicting about $6.2$ million children in the United States. However, asthma could be better managed by identifying and avoiding triggers, educating about medications and proper disease management strategies. This research utilizes deep learning methodologies to predict asthma-related emergency department (ED) visit within $3$ months using Medicaid claims data. We compare prediction results against traditional statistical classification model - penalized Lasso logistic regression, which we trained and have deployed since $2015$. The results have indicated that deep learning model Artificial Neural Networks (ANN) slightly outperforms (with AUC $= 0.845$) the Lasso logistic regression (with AUC $= 0.842$). The reason may come from the nonlinear nature of ANN.

\end{abstract}

\keywords{Pediatric Asthma \and Emergency Department Visit \and 
				 Administrative Claims Data \and Deep Learning \and 
				 Predictive Analytics \and ANN}

\section{Introduction}

More than $25$ million Americans have asthma, which affects $7.7\%$ of adults and $8.4\%$ of children \cite{CDC_gov_2018}. Asthma has been increasing since the early $1980$s in all age, sex and racial groups. Currently, asthma afflicts about $6.2$ million children under the age of $18$ in the United States \cite{zahran2018vital}, rendering it the leading chronic childhood illness. Asthma accounts for $9.8$ million doctor’s office visits, $188,968$ discharges from hospital inpatient care and $1.8$ million emergency department (ED) visits each year \cite{CDC_gov_2019}. From $2008$ to $2013$, the annual economic cost of asthma was more than $\$81.9$ billion, including $\$50.3$ billion medical costs \cite{nurmagambetov2018economic}. The median annual medical cost of asthma was $\$983$ per child in $2012$, with a range of $\$833$ in Arizona to $\$1,121$ in Michigan \cite{nurmagambetov2017state}. 
Asthma disproportionately affects low-income, minority and Medicaid-insured children, causing increased condition issues \cite{pacheco2014homes}. There is no cure to asthma, but studies reveal that it can be managed with proper disease management and adequate medical treatment, which can effectively improve asthma-related outcomes and reduce health care costs \cite{greineder1999randomized, karnick2007pediatric}.

Asthma is an ambulatory care-sensitive condition and most exacerbations that lead to ED visits or hospitalizations are avoidable \cite{das2017predicting}. Timely identification of high risk asthma patients and proactive interventions are the key to improving asthma care in the long-term. Risk factors for asthma-related adverse events have been extensively studied and multi-factorial risk predictive models are playing an increasingly recognized role in optimizing value by focusing asthma care on those at greatest risk \cite{ahmadizar2016childhood, tolomeo2009predictors}.  
These models are mostly trained as traditional statistical models, including logistic regressions, $k$-nearest meighbors, decision trees and support vector machines \cite{dreiseitl2002logistic}.

Parkland Center for Clinical Innovation (PCCI) has developed a Lasso logistic regression model in $2015$ to predict asthma ED visit in the following $3$ months for children under $18$ years old, using clinical, health services utilization and socio-demographic variables from Medicaid claims data. Compared to the published predictive models, our model has higher clinical relevance \cite{xu2011genome}, shows decent predictive accuracy \cite{forno2010risk}, is derived from relatively large populations \cite{schatz2003risk}, and is well-evaluated \cite{andrews2013asthma}. We have been sending monthly alert reports to providers, which contain information of predicted high risk asthma patients, as well as inserting the Best Practice Alert (BPA) into Epic system, aiming to reduce unnecessary hospital utilization and cost, increase patient adherence to medication and clinic visit, and improve overall health care experience.

In this research, we continued to focus on Medicaid pediatric patients from Parkland Community Health Plan (PCHP), a Medicaid health management organization (HMO) in north Texas, who provided us with the study setting. Our primary goal was to utilize administrative claims data for a Medicaid-enrolled pediatric patient population to train and test deep learning predictive models for forecasting the risk of asthma-related ED visits or hospitalizations within the next $3$ months and to evaluate if the emerging deep learning models could outperform the current Lasso logistic regression already in practice by comparing their predictive power.
Our Lasso logistic regression model served as the baseline benchmark against which deep learning model results would be compared. To the best of our knowledge, this approach for this particular use case leveraging administrative claims data has not been attempted before. 


\section{Background}

By the Council of State of Territorial Epidemiologists (CSTE) definition, a patient has ``probable" asthma if s/he had at least one ED visit or hospitalization or outpatient visit with a primary diagnosis of asthma, or at least one asthma medication prescription in the preceding $12$ months \cite{wakefield2006modifications}.
We used this CSTE standard to identify our original cohort population in this study.

Health systems and clinicians have relied on traditional reporting tools for asthma case management and risk assessment to imporve care quality. The Healthcare Effectiveness Data and Information Set (HEDIS) definition for persistent asthma is a commonly used set of criteria \cite{berger2004utility, mosen2005well, gelfand2006use}, which is a combination checking of a patient's asthma-related ED visit, hospitalization and outpatient visit, as well as asthma medication dispensing events. We refer to \cite{wakefield2006modifications} for detailed medication prescription criteria. Note that HEDIS persistent asthma definition is a stronger condition than CSTE ``probable" asthma definition.  
The asthma medication ratio (AMR) also enjoys wide usage in identifying high risk asthma patients \cite{andrews2013asthma, broder2010ratio}, which is defined as follows:
\begin{equation*}
\text{AMR} = \frac{
	\text{number of controller prescription fills}
	}{
	\text{number of controller prescription fills} + \text{number of reliver prescription fills}
	}.
\end{equation*}
An $\text{AMR} < 0.5$ is usually associated with higher risk of patients ending up in the hospital with an acute asthma exacerbation in the following several months \cite{andrews2013asthma, stanford2013predicting, schatz2006controller}. Both HEDIS and AMR criteria are typically computed from a $12$-month time cycle. And neither of them addresses the socio-demographic or comorbidities factors.

PCCI's Lasso logistic regression model is robust and clinically relevant, which significantly outperforms both HEDIS persistent asthma case-definition criteria and AMR$<0.5$ clinical criteria in the predictive power to classify high risk patients.

\section{Methodology}

Deep learning models identifies intricate structure in large data sets \cite{lecun2015deep} through multiple layers in the neural network architectures that learn directly from the data without the need for manual feature extraction.
Health care stands to benefit immensely from deep learning technologies because of the data volume, as well as the emerging unstructured complex types of data including electronic health records (EHR), imaging and text data \cite{esteva2019guide, miotto2017deep}. Recently various deep learning models have been extensively applied to different subfields in health care and have achieved great advancements \cite{reddy2018predicting, pham2016deepcare, badgeley2019deep, osmani2018processing, lauritsen2019early}.

\subsection{Data and Features}

Claims data consists of billing codes that health care providers and facilities submit to payers. It follows a consistent format and uses a standard set of pre-established codes that describe specific diagnosis, procedures, medications, as well as billed and paid amounts \cite{claims_EHR}. Additionally, claims data documents nearly all interactions a patient has across all the health care systems. Claims data captures broader information for patients and provides access to larger and more diverse patient cohort. However, claims data has, by its nature, a time lag of about $30$ to $90$ days due to the processing time before it is finally added to the database and becomes available for analysis use. 

\begin{figure}
\centering
\includegraphics[width = 0.7\textwidth]{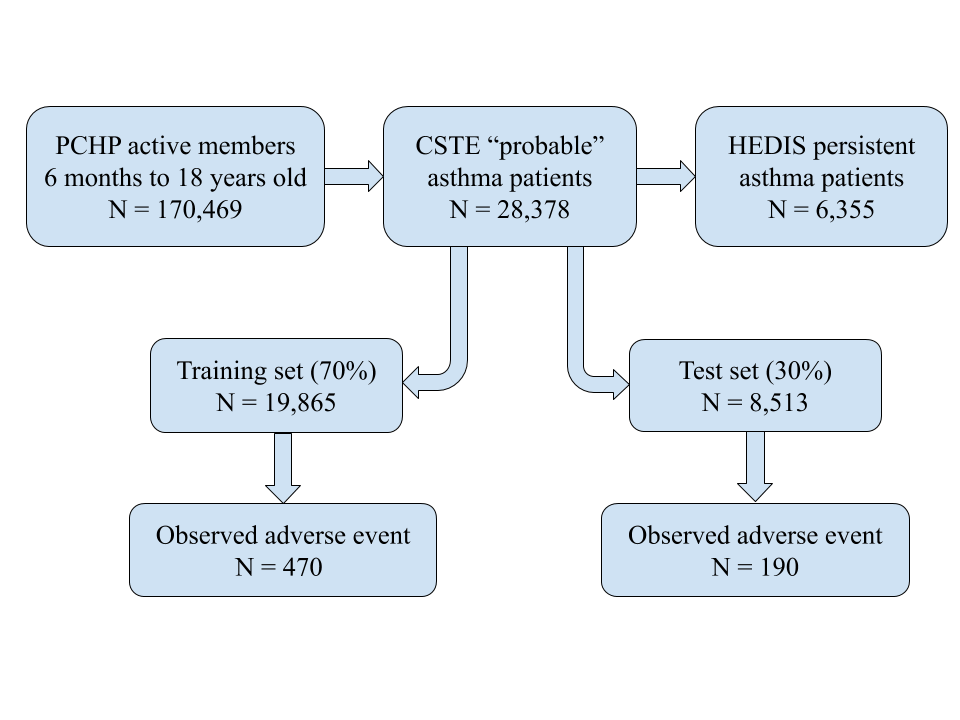}
\caption{Asthma cohort selection process}
\label{fig:cohort_selection}
\end{figure}

In order to compare with our baseline model in production, the data for this study was extracted from PCHP claims data between July $2012$ and June $2014$, which was the same time range as the one we used to train the Lasso logistic regression model. We first filtered for children aged between $6$ months and $18$ years old at prediction time and applied CSTE ``probable'' asthma criteria to identify original cohort. To be included in the model cohort, we further checked that the patient was enrolled in PCHP continually in our study time range. The size of the final data set was reduced to $28,378$ unique patients. And the prevalence rate for patients with asthma ED visits in the following $3$ months was about $3\%$. The Figure \ref{fig:cohort_selection} illustrates and summarizes the cohort selection process for our study.

The features generated from claims data for predictive model could be broadly classified into five categories, as shown in the Table \ref{table: features}. Considering the inherent $30$-day time lag in PCHP claims data, we excluded the features that were highly likely to be incorrectly coded, for example the ED visit number in the past month at prediction time. After this feature pre-selection step, we had in total $33$ features. 

\begin{table}[hbt!]
  \centering
  \begin{tabular}{c|c}
    \toprule
    Category  &  Example features    \\
    \midrule
    Demographics  &  Gender, age     \\
    Medication  &  AMR, controller medication dispensation events, reliever medication dispensation events  \\
    Health service utilization  & Number of asthma-related ED visits in the past $3, 6, 12$ months  \\
    Comorbid illinesses  &  Obesity, sleep apnea    \\
    Insurance gap & Number of insurance gaps in the past $12$ months   \\
    \bottomrule
  \end{tabular}
\label{table: features}
\caption{Model features in broader categories}
\end{table}

\subsection{Methods}

\begin{figure}
\centering
\includegraphics[width = 0.7\textwidth]{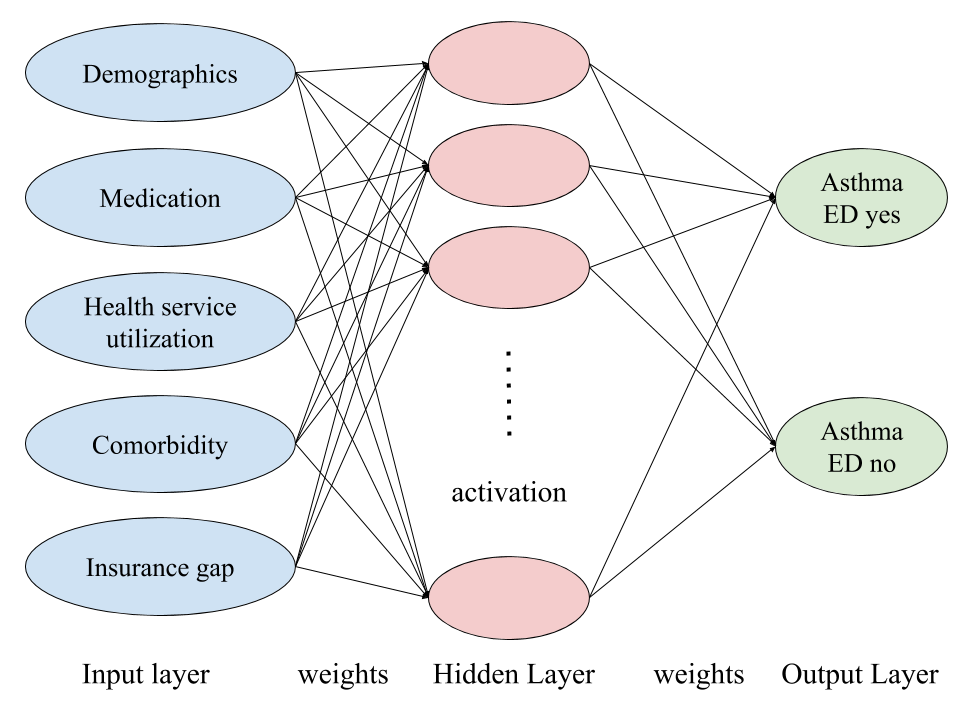}
\caption{ANN connection in our model}
\label{fig: ANN_architecture}
\end{figure}

Artificial Neural Networks are inspired by biological neural networks, which are based on a family of interconnected units, called \emph{artificial neurons}. Each connection can receive a signal from artificial neurons as an input, change the internal state and transmit the output to another artificial neurons connected to it. ANNs could learn and model complex nonlinear relationships in the data sets \cite{daniel2013principles}. The feedforward networks are used in our study and the model structure is shown in the Figure \ref{fig: ANN_architecture}. The input layer consists of all the predictors that were previously validated and normalized when necessary. The hidden layer applies the activation function to the weighted sum of input layers and forward passes the results to the next layer. In the final output layer, the \emph{sigmoid} activation function is used to produce a probability for our desired binary outcome. We define the loss function and backpropagate the error to hidden layers to update the weights. We iterate this process until the predefined convergence rate is achieved. 

\subsection{Training and Test Strategies}

Data was randomly divided into training and test sets in $0.7/0.3$ proportions. We applied various resampling methods to obtain different training samples $-$ original training data, oversampled data and downsampled data. We applied dropout regularization technique to avoid overfitting. Considering the size of the data set, we only used two hidden layers as one hidden layer are usually sufficient for most classification problems and too many hidden layers could easily cause overfitting \cite{Goodfellow-et-al-2016}. 
The model was trained using \emph{Keras} library in \emph{Python $3.7$} and the Table \ref{table: hyperparameters} shows our best hyperparameter selection of the ANN model.

\begin{table}[hbt!]
  \centering
  \begin{tabular}{c|c}
    \toprule
    Parameters  &   Selection    \\
    \midrule
    Loss function  &  Binary cross-entropy     \\
    Optimizer  &  Adam  \\
    Activation function  & Leaky ReLU ($\alpha = 0.1$) and Sigmoid  \\
    Batch size  &  $32$    \\
    Epochs & $100$ \\
    Dropout (recurrent dropout) &  $0.5$ \\
    Learning rate & $0.01$ \\
    \bottomrule
  \end{tabular}
\label{table: hyperparameters}
  \caption{Hyperparameter selection in ANN model}
\end{table}


\section{Results and Discussion}

\begin{figure}
\centering
\includegraphics[width = 0.7\textwidth]{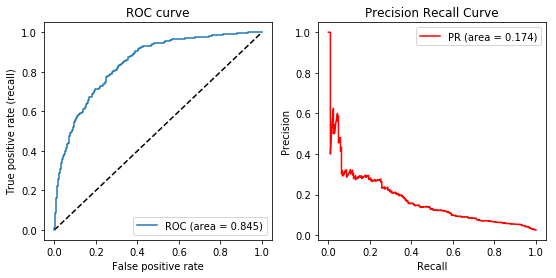}
\caption{ROC curve and PR curve of ANN}
\label{fig: ANN_ROC_prcurve}
\end{figure}

We provided the following statistical metrics from the test data to evaluate and compare the classification power between ANN model and Lasso logistic regression model: the area under the Receiver Operating Characteristic curve (ROC AUC, C-statistic), recall, precision, $F_1$ score and the area under precision-recall curve (PR AUC). Here $F_1$ score is defined as $\displaystyle F_1 = \frac{2 \cdot \text{precision} \cdot \text{recall}}{\text{precision} + \text{recall}}$. Due to the significant imbalance in the test data, we didn't utilize prediction accuracy to assess model performance. Based on the prevalence rate of actual adverse events, model performance metrics and clinical assessment of the likely capacity of a potential intervention program, we designed the patients in the top most $10$ percent of risk scores as ``High", the $10^{\text{th}}$ to $20^{\text{th}}$ percentile range as ``Medium'' and the rest as ``Low'' risk in the Lasso logistic regression model. We kept the same thresholds for ANN model. The predicted adverse events were only from ``High'' risk category. Results from proposed models are shown in the Table \ref{table: results} and the Figure \ref{fig: ANN_ROC_prcurve}. Larger areas from both ROC curve and PR curve for ANN model indicated that it outperformed the Lasso logistic regression in classification power. However, lower $F_1$ reminded us that we need to re-adjust our thresholds for ANN model.

\begin{table}[hbt!]
  \centering
  \begin{tabular}{c|ccccc}
    \toprule
    Models & ROC AUC & Recall & Precision & $F_1$ score & PR AUC \\
    \midrule
    Lasso logistic regression & $0.842$ & $0.565$ & $0.123$ & $0.202$ & $0.170$\\
    ANN & $0.845$ & $0.510$ & $0.123$ & $0.198$ & $0.174$ \\
    \bottomrule
  \end{tabular}
  \label{table: results}
  \caption{Evaluation metrics for model performance}
\end{table}

\section{Conclusion}

With the exact same data set and initial feature list, ANN model only produced slightly higher statistical classification power than the Lasso logistic regression. This is consistent with the results from \cite{dreiseitl2002logistic} to compare logistic regression and ANN models in multiple medical data classification tasks. This study further confirmed that the Lasso logistic regression model developed by PCCI in $2015$ could produce desirable statistical performance that is non-inferior to deep learning models which are more difficult to interpret. And in order for our predictive models to be deployed and effectively improve patient care, we need to work closely with clinicians to explain predictions in comprehensive and interpretable formats to build trust and transparency with stakeholders.

For future studies, blender algorithms would be tested against other singular models to achieve better statistical performances. We would explore the temporal relationships in claims data using other deep learning models, like Recurrent Neural Networks (RNN) and Long Short-Term Memory (LSTM). In order to get more timely and accurate patient information, we could link EHR data, which is longitudinal in nature, with claims data. We could also combine Social Determinants of Health (SDoH) data with claims data, to analyze how it contributes to the outcome. We would continue validating our model using the most recent available claims data and retrain the model if necessary to capture more accurate characteristics of our cohort populations.

\section{Acknowledgment}

We  acknowledge Parkland Community Health Plan (PCHP) for providing us with the data and giving us the opportunity to work on this project. We are thankful to the leadership of Parkland Health and Hospital System (PHHS) for the support.

%

\bibliographystyle{abbrv}
\bibliography{asthma_reference}

\end{document}